\documentclass{article}
\usepackage{import}
\usepackage{titling}
\usepackage{xpatch}
\makeatletter
  {\begin{titlepage}\def\@noticestring{}}%
  {\end{titlepage}}
\xpatchcmd\titlepage{\setcounter{page}\@ne}{}{}{}
\xpatchcmd\endtitlepage{\setcounter{page}\@ne}{}{}{}
\makeatother

\usepackage[final, nonatbib]{neurips_2022}

\usepackage[utf8]{inputenc} 
\usepackage[T1]{fontenc}    
\usepackage[dvipsnames]{xcolor}         
\usepackage[pagebackref=true]{hyperref}        
\hypersetup{
    colorlinks = true,
    linkcolor = BrickRed,
    citecolor = MidnightBlue,
    linktoc=all,
}
\usepackage{url}            
\usepackage{booktabs}       
\usepackage{amsfonts}       
\usepackage{nicefrac}       

\usepackage{graphicx}
\usepackage{subcaption}
\usepackage{xspace}
\usepackage{amsmath,amssymb,bm,stmaryrd}
\usepackage{multirow}
\usepackage{wrapfig}
\usepackage{tabu}
\usepackage{sidecap, caption}
\usepackage{colortbl}
\usepackage[capitalise]{cleveref}
\usepackage{algorithm}
\usepackage{algpseudocode}
\usepackage{lipsum}

\def\ie{i.e.\xspace}

\def\etal{\textit{et al.}\xspace}

\DeclareMathOperator*{\argmax}{arg\,max}
\DeclareMathOperator*{\argmin}{arg\,min} 

\newcommand{\x}{\bm{x}}
\newcommand{\z}{\bm{z}}

\newcommand{\cD}{\mathcal{D}}
\newcommand{\cX}{\mathcal{X}}
\newcommand{\cY}{\mathcal{Y}}
\newcommand{\cZ}{\mathcal{Z}}
\newcommand{\cL}{\mathcal{L}}
\newcommand{\cN}{\mathcal{N}}
\newcommand{\cG}{\mathcal{G}}
\newcommand{\E}{\mathbb{E}}

\newcommand{\Rm}{\mathbb{R}^m}

\newcommand{\pdata}{p_{\text{data}}}
\newcommand{\ptheta}{p_{\theta}}

\newcommand{\Etheta}{E_{\theta}}
\newcommand{\EMoG}{E_{\textit{\tiny{MoG}}}}
\newcommand{\EMaha}{E_{\mathcal{G}}}

\newcommand{\Lmle}{\cL_{\textit{\tiny{MLE}}}}
\newcommand{\Lreg}{\cL_{\textit{\tiny{Reg}}}}
\newcommand{\Ltot}{\cL_{\textit{\tiny{Tot}}}}

\newcommand{\minus}{\scalebox{0.65}[0.9]{$-$}}

\DeclareCaptionLabelFormat{minipagetable}{\tablename~\thetable}

\title{Energy Correction Model in the Feature Space for Out-of-Distribution Detection}
\author{%
  Marc Lafon \textsuperscript{1} \\
   \small{\texttt{marc.lafon@lecnam.net}}
   \And
  Cl{\'e}ment Rambour \textsuperscript{1} \\
  \small{\texttt{clement.rambour@cnam.fr}}
  \And 
  Nicolas Thome \textsuperscript{2} \\
  \small{\texttt{nicolas.thome@isir.upmc.fr}}
  \And\\
\textsuperscript{1}{CEDRIC, Conservatoire National des Arts et Métiers, Paris, France} \\
\textsuperscript{2}{ISIR, Sorbonne Université, Paris, France} \\
} 

\begin{document}
\begin{titlepage}
\maketitle

\begin{abstract}
In this work, we study the out-of-distribution (OOD) detection problem through the use of the feature space of a pre-trained deep classifier. We show that learning the density of in-distribution (ID) features with an energy-based models (EBM) leads to competitive detection results. However, we found that the non-mixing of MCMC sampling during the EBM's training undermines its detection performance. To overcome this an energy-based correction of a mixture of class-conditional Gaussian distributions. We obtains favorable results when compared to a strong baseline like the KNN detector on the CIFAR-10/CIFAR-100 OOD detection benchmarks.

\end{abstract}

\section[1 Introduction]{Introduction}
\label{sec:introduction}
Out-of-distribution (OOD) detection is a major safety requirement for the deployment of deep learning models in critical applications. In this work, we focus on out-of-distribution detection methods that use a pre-train classifier and we assume that OOD samples are not available neither during the training of the classifier nor to train the OOD detector. 

Lee \etal \cite{mahalanobis2018} have proposed the Mahalanobis detector which amounts to model the in-distribution (ID) features with a mixture of class-conditional Gaussian distributions (MoG). In \cite{Sehwag2021}, the authors have shown that using the Mahalanobis distance on normalized features improves performance both for a supervised and self-supervised backbone. Recently, the authors in \cite{Sun2022} have pointed out that a simple k-nearest-neighbors distance greatly improves detection performance, especially for OOD samples lying in the vicinity of ID samples. This result suggests that the Gaussian assumption may not be necessarily verified. Nonetheless, the Mahalanobis detector performs better on samples that are far-away from the ID samples. Hence, we would like to overcome the limitation of the Mahalanobis detector while conserving its strengths. 

While showing great success on several image modeling tasks \cite{ Du2019, Gao2020, Gao2021}, energy-based models (EBMs) \cite{Lecun2006} are still under-explored to estimate the density of ID samples in the feature space of a pre-trained classifier. An EBM is an unnormalized density model entirely defined by an energy function. It can be learned by maximum likelihood estimation (MLE), which consists in diminishing the energy of real samples and increasing the energy of synthesized samples. These synthesized samples are generally generated via a gradient-based Markov Chain Monte Carlo (MCMC) sampling such as Stochastic Gradient Langevin Dynamics (SGLD) \cite{WellingT11}. However, gradient-based MCMC sampling is known to struggle to sample all the modes during training, especially in high-dimensional space, thus harming density estimation.

Building upon ideas from energy-based correction literature \cite{Arbel2021, Dai2016, Gao2020, Gutmann2012, Nijkamp2022, Xie2016}, we propose to learn an energy-based model to refine a mixture of class-conditional Gaussian distributions in the feature space of a pre-trained classifier. The mixture of Gaussian distributions allows to cover all the modes and ensures a decreasing density function as we move away from the training data. On the other hand, the energy-based model will reshape the Gaussian densities, as illustrated in Figure \ref{fig:energy_toy_ebm} on a toy dataset, and will be less dependent on the mixing of MCMC sampling. 

\textbf{Contributions.} Our contributions are threefold: (\textbf{1}) we are, to the best of our knowledge, the first to show that training an EBM in the feature space leads to competitive detection performance 
(\textbf{2}) we introduce an energy-based correction model that improves both the Mahalanobis distance and the EBM, (\textbf{3}) we demonstrate favorable results on the CIFAR-$10$ and CIFAR-$100$ OOD detection benchmarks with respect to a strong baseline like the KNN detector.

\begin{figure}[t]
\centering
\begin{minipage}[c]{0.24\textwidth}
\centering
    \includegraphics[width=0.7\linewidth, height=0.7\textwidth]{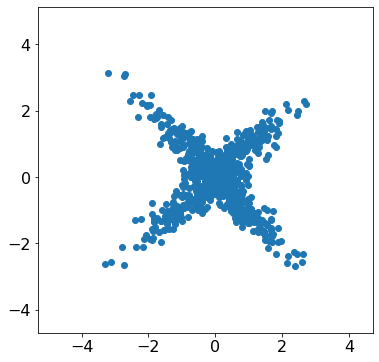}
   \subcaption{Toy dataset}
   \label{fig:toy_dataset}
\end{minipage}%
\begin{minipage}[c]{0.24\textwidth}
\centering
    \includegraphics[width=0.8\linewidth, height=0.7\textwidth]{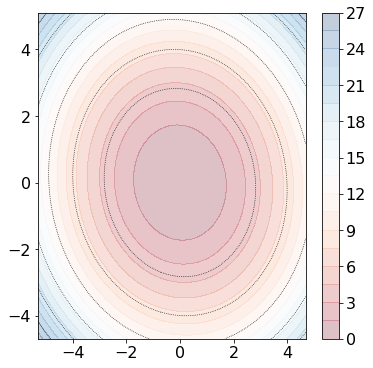}
    \subcaption{Gaussian}
    \label{fig:energy_gaussian}
\end{minipage}%
\hspace{-1mm}+\hspace{-2mm}
\begin{minipage}[c]{0.24\textwidth}
\centering
    \includegraphics[width=0.8\linewidth, height=0.7\textwidth]{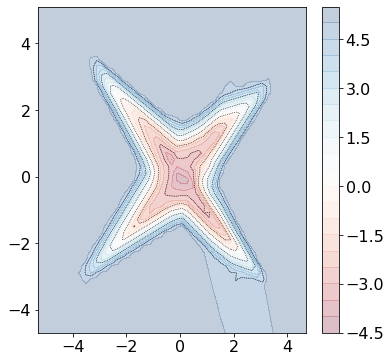}
    \subcaption{Energy correction}
    \label{fig:energy_nn}
\end{minipage}
=
\begin{minipage}[c]{0.24\textwidth}
\centering
    \includegraphics[width=0.8\linewidth, height=0.7\textwidth]{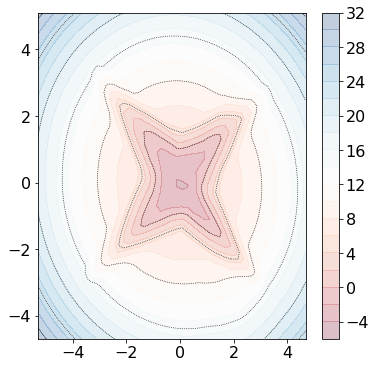}
    \subcaption{ours}
    \label{fig:energy_ours}
\end{minipage}%
\vspace{15pt}
\caption{\text{{Illustration of our energy-based correction model on a 2D toy dataset}.} The energy from the Gaussian (a) is corrected by the energy-based model (b) to produce the final energy (c)}
\label{fig:energy_toy_ebm}
\end{figure}

\section[2 Method]{Energy-based correction model in feature space}
\label{sec:Method}
We aim to estimate the density in the feature space $\cZ$ of a pre-trained deep classifier ${f = h \circ \phi}$, where ${\phi : \cX \rightarrow \cZ}$ is the feature extractor and ${h: \cZ \rightarrow \cY}$ the classification head. We suppose that we only have access to the in-distribution dataset ${\cD=\{(\x_i,y_i)\}_{i=1}^N}$ used to train the classifier $f$. 

\subsection[2.1 Model]{Model}
\label{sec:model}
We would like to combine in one model the performance of the Mahalanobis detector on OOD examples that are far away from the ID data, and the flexibility of energy-based models to improve detection of closer OOD samples. To this end, we introduce an energy-based correction of a reference distribution ${q(\z)}$ which is a mixture of class-conditional Gaussian distributions:
\begin{equation}
\label{eq:model}
\ptheta(\z) \propto \exp{(\minus\Etheta(\z))}q(\z),
\quad \text{with}\quad q(\z) = \sideset{}{_c}\sum \pi_c \cN(\z;\mu_c,\Sigma),
\end{equation}
where the energy-correction $\Etheta$ is a neural network with parameter $\theta$, and where we use the empirical estimates for the parameters of the mixture and assume a tied covariance matrix, like in \cite{mahalanobis2018} (more details in appendix \ref{supp:sec:MoG}). In the sequel, we will write ${q(\z) \propto \exp{(\minus\EMaha(\z))}}$ with ${\EMaha(\z)=\minus \log\sum_c \exp{(\minus(\z\minus\mu_c)^T\Sigma^{\minus1}(\z\minus\mu_c))}}$ so that $\ptheta$ can be expressed as ${\ptheta(\z) \propto \exp{(\minus(\Etheta(\z) + \EMaha(\z)))}}$.

\subsection[2.2 Training]{Training}
We train our correction model via maximum likelihood estimation (MLE) which amounts to perform stochastic gradient descent with the following loss\footnote{see more details on EBMs training in the appendix}:
\begin{equation}
\label{eq:CD}
\Lmle =\E_{\z \sim p_{in} }\big[\Etheta(\z)\big] - \E_{\z' \sim \ptheta}\big[\Etheta(\z')\big],
\end{equation}
where we write $\z \sim p_{in}$ for $\z = \phi(\x)$ with $\x \sim \pdata.$ Note that the energy inside the expectations in (\ref{eq:CD}) should be $\Etheta + \EMaha$, but we discarded $\EMaha$ as it does not depend on $\theta$.

\textbf{SGLD sampling.}
To compute $\Lmle$ we must sample synthetic features with respect to our current model $\ptheta$.  We follow previous works \cite{Du2019, Grathwohl2019} and use stochastic gradient Langevin dynamics (SGLD) sampling \cite{WellingT11}. An SGLD iteration step writes
\begin{equation}
\label{eq:SGLD}
\z_0 \sim q,\quad \quad \z_{t+1} = \z_{t} - \alpha_t \nabla_{\z}(\Etheta+ \EMaha)(\z_t) + \sqrt{\beta_t} \epsilon, \quad \quad \epsilon \sim \cN(0,I),
\end{equation}
where $\alpha_t$ is the step size and $\beta_t$ is the noise scale.

Because we would like to rely on $\EMaha$ as we move away from ID samples, we initiate the SGLD samples with the reference distribution $q$ instead of a normal or uniform distribution as usually done in the EBM literature. This way, $\Etheta$ will refine $\EMaha$ only on high density areas. We stress the importance of taking gradient steps with respect to $\Etheta+\EMaha$ in equation (\ref{eq:SGLD}), contrarily with equation (\ref{eq:CD}) where we could ignore $\EMaha$.

\textbf{$\boldsymbol{L_2}$-regularization.}
Without any modification to the loss, $\Etheta$ could dominate $\EMaha$ which is an undesirable behavior. To avoid this, we add an $L2$-regularization on the magnitudes of $\Etheta$ which also helps stabilize training. The final loss is then $\ \Ltot =\Lmle + \alpha\ \Lreg,\ $ where $\alpha$ is an hyper-parameter controlling the strength of the $L_2$-regularization.

\section[3 Results]{Results}
\label{sec:Experiments}

In this section, we first present visual results on a 2D synthetic example to give a better grasp of our approach. Then, we compare our approach against state-of-the-art baselines for OOD detection on image datasets and perform ablation experiments.

\subsection[3.1 Synthetic datasets]{Synthetic datasets}
\label{sec:toy_dataset}
\vspace{-2mm}
\begin{wrapfigure}{r}{0.45\linewidth}
\centering
\begin{minipage}[c]{0.24\textwidth}
\hspace{-5mm}
\centering
    \includegraphics[width=0.7\linewidth, height=0.7\textwidth]{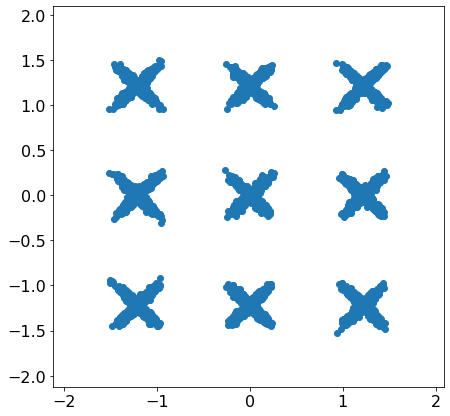}
   \vspace{-2mm}
   \subcaption{\small{Dataset}}
   \label{fig:toy_mix_dataset}
\end{minipage}
\hspace{-6mm}
\begin{minipage}[c]{0.24\textwidth}
\centering
    \includegraphics[width=0.8\linewidth, height=0.7\textwidth]{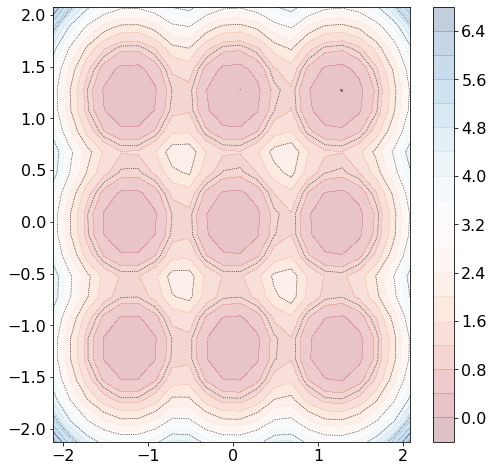}
    \vspace{-2mm}
    \subcaption{\small{MoG}}
    \label{fig:toy_mix_energy_gaussian}
\end{minipage}\\[10pt]
\begin{minipage}[c]{0.24\textwidth}
\centering
    \includegraphics[width=0.8\linewidth, height=0.7\textwidth]{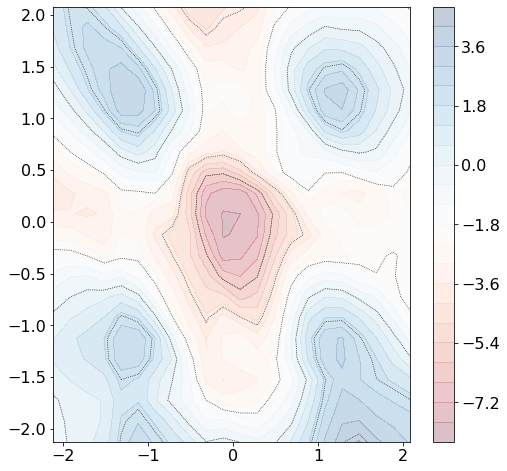}
    \vspace{-2mm}
    \subcaption{\small{EBM}}
    \label{fig:toy_mix_energy_nn}
\end{minipage}%
\hspace{-6mm}
\begin{minipage}[c]{0.24\textwidth}
\centering
    \includegraphics[width=0.8\linewidth, height=0.7\textwidth]{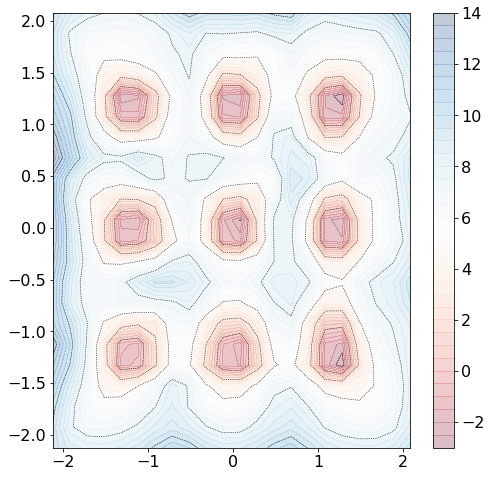}
    \vspace{-2mm}
    \subcaption{\small{ours}}
    \label{fig:toy_mix_energy_ours}
\end{minipage}%
\vspace{5mm}
\caption[]
    {\tabular[t]{@{}l@{}}EBM's energy is biased due\\ 
    \hspace{2mm} to SGLD non-mixing\endtabular}
\label{fig:energy_toy_mixing}
\vspace{6mm}
\end{wrapfigure}

First, we design a simple 2D synthetic dataset consisting of data points sampled from a "crossed" distribution (Fig. \ref{fig:toy_dataset}). In Figure \ref{fig:energy_toy_ebm}, we can visualize the result of our proposed approach on this dataset. The final energy of our model is the sum of the Gaussian energy and of the neural network's energy which learns a residual. We can see in Figure \ref{fig:energy_ours} that our learned energy is sharp nearby the training data and becomes smoother as we move away from the data.

The second synthetic experiment is designed to illustrate that EBMs tend to learn a biased energy function when the MCMC is not mixing, especially when there are multiple modes in the target distribution. We consider a dataset consisting of 9 cross datasets distributed according to a 3 $\times$ 3 grid (Fig. \ref{fig:toy_mix_dataset}). We can see on Figure \ref{fig:toy_mix_energy_nn} that the EBM struggles to cover all the modes while the mixture of Gaussian (Fig. \ref{fig:toy_mix_energy_gaussian}) and ours (Fig. \ref{fig:toy_mix_energy_ours}) both perfectly recover all the modes.

\subsection[3.2 Image datasets]{Image datasets}
 \label{sec:imageexp}
\textbf{Baselines.} We evaluate our approach against the following baselines: the maximum softmax probability (MSP) \cite{hendrycks17baseline}, ODIN \cite{odin2018}, and Energy-logits \cite{Liu2020}. We also compare against methods that aim to estimate the density of in-distribution samples in the feature space: SSD \cite{Sehwag2021} and KNN \cite{Sun2022}. Since they obtain better performances with normalized features (\ie $\z = \phi(\x) / \lVert \phi(\x) \rVert$), we use normalize features for all such methods (including ours).

\textbf{Evaluation metrics.} We report the following standard metrics used in the literature \cite{hendrycks17baseline}: the area under the receiver operating characteristic curve (AUC) and the false positive rate at a threshold corresponding to a true positive rate of 95\% (FPR95).

\textbf{Datasets.} We conduct experiments using CIFAR-$10$ and CIFAR-$100$ datasets \cite{krizhevsky2009learning} as in-distribution datasets. For OOD datasets, we define three categories: near-OOD datasets, mid-OOD datasets and far-OOD datasets. These correspond to different levels of proximity with the ID datasets. For CIFAR-$10$ (resp. CIFAR-$100$), we consider \texttt{TinyImageNet}\footnote{https://www.kaggle.com/c/tiny-imagenet} and CIFAR-$100$ (resp. CIFAR-$10$) as near-OOD datasets. Then for both CIFAR-$10$ and CIFAR-$100$, we use \texttt{LSUN} \cite{yu15lsun} and \texttt{Places365} \cite{zhou2017places} datasets as mid-OOD datasets, and \texttt{Textures} \cite{cimpoi14describing} and \texttt{SVHN} \cite{svhn-dataset} as far-OOD datasets.

\textbf{Implementation details.} All experiments were conducted using \texttt{PyTorch}. We use a ResNet-34 classifier from the \texttt{timm} library \cite{rw2019timm} for both ID datasets. The classifier is trained for $200$ epochs using SGD with momentum and learning rate $0.1$. Our energy correction model consists in a $6$ layers MLP trained for $20$ epochs with Adam with learning rate $5e\text{-}6$ (all training details are available in the appendix).

\begin{table}[t]
\caption{\textbf{Results on CIFAR-10 \& CIFAR-100.} All methods are based on a pre-trained ResNet-34 trained on the ID dataset only. $\uparrow$ indicates larger is better and $\downarrow$ the opposite. Best results are in bold.}
\centering
\begin{subtable}[c]{\textwidth}
\centering
	\resizebox{0.85\linewidth}{!}{%
    	\begin{tabular}{l  c  c  c  c}
    		\toprule
        &  \textit{Near-OOD} & \textit{Mid-OOD} & \textit{Far-OOD} & \textbf{Average}\\
        \multirow{-2}{*}{\textbf{Method}} &  \small{\text{FPR95$\downarrow$ / AUC $\uparrow$}} & \small{\text{FPR95$\downarrow$ / AUC $\uparrow$}}  & \small{\text{FPR95$\downarrow$ / AUC $\uparrow$}}  & \small{\text{FPR95$\downarrow$ / AUC $\uparrow$}}\\
    		\midrule
    		MSP &  57.0 / 88.0 & 51.6 / 91.0 &  36.0 / 94.4 & 48.2 / 91.2\\ 
    		ODIN & \underline{45.3} / 86.7 & \textbf{34.1} / 91.3 & 18.6 / 95.5 & \textbf{32.6} / 91.1\\
                Energy-Logits  &  \underline{45.2} / 87.6 & \textbf{34.7} / 91.8 & 17.8 / 96.0 & \textbf{32.6} / 91.8\\ 
                KNN & \textbf{42.3} / \textbf{90.5} &  \underline{37.0} / \textbf{93.4} &  16.7 / \underline{97.3} & \textbf{32.9} / \textbf{93.7}\\ 
                SSD &  51.8 / 89.3 &  46.8 / 91.8 &  \textbf{7.0 / 98.8} & 35.1 / \textbf{93.3}\\ 
                \rowcolor[gray]{.95} \textbf{ours} &  47.5 / \underline{89.9} & 41.0 / \underline{92.5} &  \underline{8.7} / \textbf{98.6}&  \textbf{32.4 / 93.7}\\
    		\bottomrule
    	\end{tabular}
    }
    \label{tab:ood_results_c10}
    \vspace{-2mm}
    \subcaption{CIFAR-$10$}
\end{subtable}
\begin{subtable}[c]{\textwidth}
    \centering
	\resizebox{0.85\linewidth}{!}{%
    	\begin{tabular}{l  c  c  c  c}
    		\toprule
        &  \textit{Near-OOD} & \textit{Mid-OOD} & \textit{Far-OOD} & \textbf{Average}\\
        \multirow{-2}{*}{\textbf{Method}} &  \small{\text{FPR95$\downarrow$ / AUC $\uparrow$}} & \small{\text{FPR95$\downarrow$ / AUC $\uparrow$}}  & \small{\text{FPR95$\downarrow$ / AUC $\uparrow$}}  & \small{\text{FPR95$\downarrow$ / AUC $\uparrow$}}\\
    		\midrule
    		MSP& \underline{79.2} / \underline{77.1} & \textbf{82.3} / \textbf{75.6} & 67.1 / 83.7 & 76.1 / 78.8\\ 
    		ODIN  & 80.1 / 76.3 & \underline{84.4} / 73.3& 71.6 / 82.8 & 78.6 / 77.5 \\ 
                Energy-Logits & 80.0 / 76.7 & 85.4 / 73.2 & 57.7 / 87.1 &  74.3 / 79.0 \\ 
                KNN  & \textbf{78.5} / \textbf{78.6} & 85.8 / \textbf{75.7} & 47.7 / 90.8 &  70.7 / \textbf{82.0} \\ 
                SSD & 84.2 / 75.3 & 86.2 / 74.2 & \underline{28.3} / \textbf{94.4} &  \underline{66.2} / \underline{81.3} \\ 
                \rowcolor[gray]{.95} \textbf{ours} & 82.9 / 76.3 &  85.2 / \underline{74.8}&  \textbf{27.0} / \textbf{94.6}&  \textbf{65.0 / 81.9}\\
    		\bottomrule
    	\end{tabular}
    }
    \label{tab:ood_results_c100}
        \vspace{-2mm}
    \subcaption{CIFAR-$100$}
\end{subtable}
\label{tab:ood_results}
\vspace{-10mm}
\end{table}

\textbf{Improved averaged OOD detection performances.}
We can see in Table \ref{tab:ood_results} that we achieve the best average results on the CIFAR-$100$ benchmark and obtain average results which are comparable to the strong baseline KNN on the CIFAR-$10$ benchmark. As expected, our approach and SSD behave similarly on far-OOD datasets, but we improve over SSD on near and mid-OOD datasets. For instance, for mid-OOD datasets, we reduces the FPR95 by $12.4\%$ on CIFAR-$10$ and increases the AUC from 91.8$\%$ to $92.8\%$ on CIFAR-$10$. On CIFAR-$100$, we improve only marginally upon SSD and remains worse than KNN on near-OOD datasets and achieves comparable results on mid-OOD.

\begin{wraptable}{r}{0.5\linewidth}
\vspace{6mm}
\caption{Ablation on CIFAR-$10$ \& CIFAR-$100$}
\label{tab:ablation}
\begin{subtable}[c]{0.5\textwidth}
\centering 
\resizebox{\linewidth}{!}{%
    \begin{tabular}{l  c  c  c  c}
        \toprule
    &  \textit{Near-OOD} & \textit{Mid-OOD} & \textit{Far-OOD} & \textbf{Average}\\
    \textbf{Method} &  \tiny{\text{FPR95$\downarrow$ / AUC $\uparrow$}} & \tiny{\text{FPR95$\downarrow$ / AUC $\uparrow$}}  & \tiny{\text{FPR95$\downarrow$ / AUC $\uparrow$}}  & \tiny{\text{FPR95$\downarrow$ / AUC $\uparrow$}}\\
        \midrule
            ours w/o energy &  51.8 / 89.3 &  46.8 / 91.8 &  \textbf{7.0 / 98.8} & 35.1 / 93.3\\ 
        \rowcolor[gray]{1.} ours w/o MoG & \textbf{44.5 / 90.2} &  \textbf{33.5} / \textbf{93.2}&  15.7 / 97.3&  31.3 / 93.5\\
            \rowcolor[gray]{1.} ours & 47.5 / \underline{89.9} & 41.0 / \underline{92.5} &  \underline{8.7} / \textbf{98.6}&  \underline{32.4} / \textbf{93.7}\\
        \bottomrule
    \end{tabular}
}
\vspace{2.5mm} 
\label{tab:ablation_c10}
\subcaption{\small{CIFAR-$10$}}
\end{subtable}
\begin{subtable}[c]{0.5\textwidth}
\centering
\resizebox{\linewidth}{!}{%
    \begin{tabular}{l  c  c  c  c}
        \toprule
    &  \textit{Near-OOD} & \textit{Mid-OOD} & \textit{Far-OOD} & \textbf{Average}\\
    \textbf{Method} &  \tiny{\text{FPR95$\downarrow$ / AUC $\uparrow$}} & \tiny{\text{FPR95$\downarrow$ / AUC $\uparrow$}}  & \tiny{\text{FPR95$\downarrow$ / AUC $\uparrow$}}  & \tiny{\text{FPR95$\downarrow$ / AUC $\uparrow$}}\\
        \midrule
            ours w/o energy & 84.2 / 75.3 & 86.2 / 74.2 & \textbf{28.3 / 94.4} &  66.2 / 81.3 \\ 
            \rowcolor[gray]{1.}ours w/o MoG & \textbf{80.9} / 75.7&  85.3 / 73.4&  63.0 / 87.6&  76.4 / 78.9\\
            \rowcolor[gray]{1.}ours  & 82.9 / \textbf{76.3} &  \textbf{85.2} / \textbf{74.8}&  \textbf{27.0} / \textbf{94.6}&  \textbf{65.0 / 81.9}\\
        \bottomrule
    \end{tabular}
 }
\vspace{2.5mm} 
\subcaption{\small{CIFAR-$100$}}
\end{subtable}\\[6pt]
\end{wraptable}

\textbf{Ablation.} To assess the importance of our energy-correction approach, we drop the reference measure $q$ in equation (\ref{eq:model}) and initiate the SGLD samples in (\ref{eq:SGLD}) with a random normal distribution $z_0 \sim \cN(0,I)$. In that case we recover a standard EBM trained with MLE via MCMC sampling. From Table \ref{tab:ablation}, we can see that even without the reference measure, the EBM achieves competitive performance with respect to previous methods on the CIFAR-$10$ benchmark. However, The EBM struggles to detect far-OOD samples on the CIFAR-$100$ benchmark. This confirms the insights we had drawn in section \ref{sec:toy_dataset} as well as our design choices.

\section[4 Conclusion]{Conclusion}
\label{sec:conclusion}
We introduce an energy-based correction model trained solely on in-distribution features of a pre-trained classifier for out-of-distribution detection. We showed that our approach improves or is on par with strong baselines like SSD of KNN on the CIFAR-$10$ and CIFAR-$100$ OOD detection benchmarks. In future work, we plan to evaluate in a low data regime scenario, with different classifier backbones, on more challenging images datasets like ImageNet \cite{DengDSLL009} and/or to extend it to tasks beyond image classification.

\newpage
\bibliographystyle{plain}
\bibliography{references}
\end{titlepage}

\appendix
\setcounter{section}{0}
\section[A Appendix]{Appendix}

\subsection[A.1 Energy-based models]{Energy-based models}
An energy-based model (EBM) is an unnormalized density model defined via its energy function $E_{\theta}:\mathbb{R}^m \rightarrow \mathbb{R}$ parameterized by a neural network with parameters $\theta$. For $\z \in \Rm$, its probability density is given by the Boltzmann distribution 

\begin{equation}
\label{eq:EBM}
p_{\theta}(\z) = \frac{1}{Z_{\theta}} \exp{(\minus\Etheta(\z)}),
\end{equation}

where $Z_{\theta}$ is the partition function which is intractable in high dimension. We can train EBMs via maximum likelihood estimation:
\begin{align}
\argmax_{\theta} \log{p_{\theta}(\mathcal{D})}=  \argmin_{\theta}\mathbb{E}_{\z\sim p_{in}}[-\log{p_{\theta}(\z)}]
\end{align}
which can be approximated via stochastic gradient descent :
\begin{equation}
\theta_{i+1} = \theta_{i} - \lambda \nabla_{\theta}( \minus\log{ p_{\theta_{i}}(\z)}) \quad \text{with}\quad \z \sim p_{in}
\end{equation}

Interestingly, $\nabla_{\theta}( \minus\log{ p_{\theta_{i}}(\z)})$ can be computed without computing the intractable normalization constant $Z_{\theta}$.

We have
\begin{align*}
 \nabla_{\theta}( \minus\log{ p_{\theta}}(\z)) &= \nabla_{\theta} E_{\theta}(\z) + \nabla_{\theta} \log{ Z_{\theta}}\\
 &= \nabla_{\theta} E_{\theta}(\z) +  \frac{1}{Z_{\theta}} \nabla_{\theta} Z_{\theta} \\
 &= \nabla_{\theta} E_{\theta}(\z)  +\frac{1}{Z_{\theta}} 
\nabla_{\theta}\int_{\z} \exp(-E_{\theta}(\z)) d\z\\
 &= \nabla_{\theta} E_{\theta}(\z) + \frac{1}{Z_{\theta}} \int_{\z} 
 \nabla_{\theta}\exp(-E_{\theta}(\z)) d\z\\
 &= \nabla_{\theta} E_{\theta}(\z) + \int_{\z}
 -\nabla_{\theta}E_{\theta}(\z) \frac{\exp(-E_{\theta}(\z)) }{Z_{\theta}} d\z\\
 &= \nabla_{\theta} E_{\theta}(\z) - \E_{\z' \sim p_{\theta}} 
 [\nabla_{\theta} E_{\theta}(\z')].
\end{align*}

Therefore, training EBMs via maximum likelihood estimation (MLE) amounts to perform stochastic gradient descent with the following loss:
\begin{equation}
\label{eq:CD2}
\Lmle =\E_{\z \sim p_{in} }\big[\Etheta(\z)\big] - \E_{\z' \sim \ptheta}\big[\Etheta(\z')\big].
\end{equation}

Intuitively, this loss amounts to diminishing the energy for samples from the true data distribution $p(x)$ and to increasing the energy for synthesized examples sampled according from the current model. Eventually, the gradients of the energy function will be equivalent for samples from the model and the true data distribution and the loss term will be zero.
 
The expectation $\E_{\z' \sim \ptheta}\big[\Etheta(\z')\big]$ can be approximated through MCMC sampling, but we need to sample $z'$ from the model $p_{\theta}$ which is an unknown moving density. To estimate the expectation under $\ptheta$ in the right hand-side of equation (\ref{eq:CD2}) we must sample according to the energy-based model $\ptheta$. To generate synthesized examples from $\ptheta$, we can use gradient-based MCMC sampling such as Stochastic Gradient Langevin Dynamics (SGLD) \cite{WellingT11} or Hamiltonian Monte Carlo (HMC) \cite{Neal2011}. In this work, we use SGLD sampling following \cite{Du2019, Grathwohl2019}. In SGLD, initial features are sampled from a proposal distribution $p_0$ and are updated for $T$ steps with the following iterative rule:
\begin{equation}
\label{eq:SGLD2}
\z_0 \sim p_0,\quad \quad \z_{t+1} = \z_{t} - \alpha_t \nabla_{\z} \Etheta(\z_{t}) + \sqrt{\beta_t} \epsilon, \quad \quad \epsilon \sim \cN(0,I),
\end{equation}
where $\alpha_t$ is the step size and $\beta_t$ the noise scale. Therefore sampling from $\ptheta$ does not require to compute the normalization constant $Z_{\theta}$ either.

Many variants of this training procedure have been proposed including Contrastive Divergence (CD) \cite{Hinton2002} where $p_0 = \pdata$, or Persistent Contrastive Divergence (PCD) \cite{Tieleman2008} which uses a buffer to extend the length of the MCMC chains. We refer the reader to \cite{Song2021} for more details on EBM training with MLE as well as other alternative training strategies (score-matching, noise contrastive estimation, Stein discrepancy minimization, etc.).

\subsection[A.2 OOD detection with EBMs]{OOD detection with energy-based models}

Once trained on in-distribution features we use the learned energy as an uncertainty score to detect out-of-distribution samples. Given an input sample $\x^*$, we compute its feature representation ${\z^* = \phi(\x^*)}$ and the decision function for out-of-distribution detection is given by ${G(\z^*) = \mathbf{1}\{\Etheta(\z^*) + \EMoG(\z^*) \geq \gamma \}}$, where $\gamma$ is a threshold which can be chosen so that at least 95 \% of the in-distribution examples are correctly classified.

\subsection[A.3 Mixture of Gaussian]{Mixture of Gaussian component details}
\label{supp:sec:MoG}

In section \ref{sec:model} of the main paper, we introduced our model as an energy-based correction of a mixture of class-conditional Gaussian distributions. Here, we provide additional details on how we compute its parameters. The mixture writes
\begin{equation}
q(\z) = \sum_c \pi_c \cN(\z; \mu_c,\Sigma),
\end{equation}

with $\pi_c$ the mixing coefficient, and $\mu_c$ and $\Sigma$ are the means and the tied covariance matrix of the Gaussian distributions. All these parameters are estimated using in-distribution features:
\begin{equation}
\label{eq:MoG}
\pi_c = \frac{N_c}{N} ,\quad \mu_c = \frac{1}{N_c}\sum_{i:y_i=c} \z_i,\quad \Sigma = \frac{1}{N} \sum_c \sum_{i:y_i=c} (\z_i-\mu_c)(\z_i-\mu_c)^T,
\end{equation}

where $N_c$ is the number of training samples of class $c$ and $\z_i = \phi(\x_i)$ for $\x_i \in \cD$.

\subsection[A.4 Baselines details]{More details about baselines}

In this section we give more details about the baseline methods we used for the comparative experiment on the image datasets (sec. \ref{sec:imageexp} of the main paper).

\paragraph{MSP.} Hendrycks \etal \cite{hendrycks17baseline} have proposed to use the maximum softmax probability of the classifier as a baseline to detect OOD detection.

\paragraph{ODIN.} \cite{Liang2018} is a threshold-based OOD detector enhancing the MSP detector with temperature scaling and inverse adversarial perturbation. Both techniques aimed to increase in-distribution MSP higher than out-distribution MSP. Temperature scaling changes the softmax probabilities using a temperature $T > 0$:
\begin{equation}
S(\x;T) = \max_c \frac{\exp(f_c( \x, \theta) / T)}{\sum_{k=1}^K \exp(f_k( \x, \theta) / T)} 
\end{equation}

and input preprocessing consists in adding a small adversarial perturbation to samples:
\begin{equation}
\tilde{\x} = \x - \epsilon~ \mathrm{sign}( \minus \nabla_{\x} \log S(\x;T))
\end{equation}

\paragraph{SSD.} Lee \etal \cite{Lee2018} have proposed to fit class-conditional Gaussian distributions with a tied covariance matrix on the penultimate layer of the neural network classifier, and to compute the maximum Mahalanobis distance to each Gaussian center as the anomaly score:
\begin{equation}
M(\x)= \min_c{-(\phi(\x) - \mu_c)^T \Sigma^{-1}(\phi(\x) - \mu_c)}.
\end{equation}

In this work we implemented the SSD \cite{Sehwag2021} score which is the Mahalanobis distance on normalized features.

In the spirit of ODIN, the authors in \cite{Lee2018} also use adversarial perturbation with the Mahalanobis score:
\begin{equation}
\tilde{\x} = \x + \epsilon~ \mathrm{sign}(\nabla_{\x} M(\x))
\end{equation}
where $\epsilon$ is the strength of the perturbation and chosen to separate ID samples from OOD samples or negative samples generated by FGSM \cite{GoodfellowSS14}. The authors have also reported improved detection performance using feature ensembling to combine the anomaly scores computed at several layers of the classifier into one unique score.  We let the exploration of these techniques for future work.

\paragraph{KNN.} Sun \etal \cite{Sun2022} proposed non-parametric density estimation using nearest neighbors in the feature space of a pre-trained classifier for OOD detection. For test image $\x^*$, they compute its feature representation $\z^*$ and then the Euclidean distances between $\z^*$ and each feature of the training dataset. The uncertainty score they use for OOD detection is the negative $k^{th}$ distance, with $k$ an hyper-parameter.

\paragraph{Feature normalization.} In the OOD detection with KNN paper  \cite{Sun2022}, the author showed that feature normalization is critical for good performance. Similarly, \cite{Sehwag2021} used normalized features for the Mahalanobis score. In the case of \cite{Sehwag2021}, the feature normalization was a byproduct of the self-supervised contrastive learning framework of their work and was not mentioned as a essential element. Their implementation using the supervised loss also used feature normalization. In this work, all results for Mahalanobis, KNN and our model are obtained with normalized feature, that is when we write $z$ we assume $z= \phi(\x) / \lVert \phi(\x) \rVert$. 

\subsection[A.5 Experimental setup]{Experimental setup}

\subsubsection[A.5.1 Training details]{Training details} In this section we provide more training details.
All experiments were conducted using the \texttt{PyTorch} library.

\paragraph{ID classifier training}
We use a ResNet-34 classifier from the \texttt{timm} library \cite{rw2019timm} for both CIFAR-$10$ and CIFAR-$100$ datasets. The classifier is trained for $200$ epochs using SGD with Nesterov momentum and weight decay. The momentum factor is 0.9 and the weight decay coefficient is $5e\text{-}4$. The learning rate is initialized at 0.1 and reduced by a factor of 10 at 50\% and 75\% of training. We use random resized crops and random horizontal flips with default parameters on images as a form of data augmentation. 

\paragraph{Training details}
Our energy-correction model consists in an MLP with $4$ hidden layers trained for $20$ epochs with Adam with learning rate $5e\text{-}6$. The network input dimension is $512$ (which is the dimension of the penultimate layer of ResNet-34), the hidden dimension is $1024$ and the output dimension is $1$. For SGLD sampling, we use 20 steps with an initial step size of $1e\text{-}6$ linearly decayed to $1e\text{-}7$ and an initial noise scale of $1e\text{-}3$ linearly decayed to $1e\text{-}4$. We add a small Gaussian noise with std $1e\text{-}3$ to each input of the network to stabilize training as done in previous work. The $L_2$ coefficient is set to $10$. We use temperature scaling on the mixture of Gaussian distributions energy with temperature $T_{\cG} = 1e3$. The hyperparameters for the CIFAR-$10$ and CIFAR-$100$ models are identical.

\paragraph{EBM training}
For the EBM we conserve the same network architecture than for our energy-correction model. We trained for $20$ epochs with Adam with learning rate $5e\text{-}5$. For SGLD sampling, we use 200 steps with an initial step size of $1e\text{-}2$ linearly decayed to $1e\text{-}3$ and an initial noise scale of $1e\text{-}2$ linearly decayed to $1e\text{-}3$. We add a small Gaussian noise with std $1e\text{-}3$ to each input of the network to stabilize training as done in previous work. We use temperature scaling on neural network's energy with temperature $T = 1e\text{-}2$. The $L_2$ coefficient is set to $0.1$. The hyperparameters for the CIFAR-$10$ and CIFAR-$100$ models are identical.

\subsubsection[A.5.2 Datasets]{Datasets}

\paragraph{ID datasets} In section \ref{sec:imageexp} of the main paper, the experiments are conducted using CIFAR-10 and CIFAR-100 datasets \cite{krizhevsky2009learning}. They consist in 32 $\times$ 32 natural images with 10 classes for CIFAR-10 and 100 classes for CIFAR-100. Both
datasets contain 50,000 training images and 10,000 test images.

\paragraph{OOD datasets} We consider the following OOD datasets: \texttt{TinyImageNet}, \texttt{LSUN} \cite{yu15lsun}, \texttt{Places} \cite{zhou2017places}, \texttt{Textures} \cite{cimpoi14describing} and \texttt{SVHN} \cite{svhn-dataset}. We use the test set of each of the previous dataset for OOD detection. \texttt{TinyImageNet} is a subset of ImageNet containing images of 200 different classes, \texttt{LSUN} contains images from 10 scene categories, \texttt{Places365} contains images from 365 scene categories, \texttt{Textures} contains images representing 47 different texture categories and \texttt{SVHN} contains images representing street view house numbers (10 categories).

\subsection[A.6 Detailed results]{Detailed results} In this section we provide detailed results for each OOD datasets in Table \ref{supp:fig:ood_results_cifar10} and Table \ref{supp:fig:ood_results_cifar100}. 

\begin{table}[ht]
\huge
\vspace{1cm}
    \centering
	\resizebox{\linewidth}{!}{%
    	\begin{tabular}{l  c c  c c  c c  c}
    		\toprule
       &  \multicolumn{2}{c}{\textit{Near-OOD}} & \multicolumn{2}{c}{\textit{Mid-OOD}}  & \multicolumn{2}{c}{\textit{Far-OOD}} & \\
  		     \textbf{Method} & \textbf{C-100} & \textbf{TinyIN} & \textbf{LSUN} & \textbf{Places365} & \textbf{Textures} & \textbf{SVHN} & \textbf{Average}\\
        &  \Large{\text{FPR95$\downarrow$ / AUC $\uparrow$}} &\Large{\text{FPR95$\downarrow$ / AUC $\uparrow$}} & \Large{\text{FPR95$\downarrow$ / AUC $\uparrow$}} & \Large{\text{FPR95$\downarrow$ / AUC $\uparrow$}} & \Large{\text{FPR95$\downarrow$ / AUC $\uparrow$}} & \Large{\text{FPR95$\downarrow$ / AUC $\uparrow$}} & \Large{\text{FPR95$\downarrow$ / AUC $\uparrow$}}\\
    		\midrule
                MSP \cite{hendrycks17baseline}&  58.0 / 87.9 & 55.9 / 88.2  & 50.5 / 91.9 & 52.7 / 90.2 &  52.3 / 91.7 & 19.7 / 97.0 & 48.2 / 91.2\\
    		ODIN \cite{odin2018} & \underline{48.4} / 86.0 & \underline{42.2} / 87.3 & \textbf{32.6} / 92.3 & \underline{35.6} / 90.4 & 29.4 / 92.6 & 7.8 / 98.3 &  \underline{32.6} / 91.1\\
                Energy-Logits \cite{Liu2020} &  \underline{48.4} / 86.9 & \textbf{41.9} / 88.2 & \underline{33.7} / 92.6 & \underline{35.7} / 91.0 &  30.7 / 92.9 & 4.9 / 99.0 &  \underline{32.6} / 91.8\\
                KNN \cite{Sun2022} & \textbf{47.9} / \textbf{90.3} & \underline{42.6} / \textbf{90.6} &  36.1 / \textbf{94.1} & 37.8 / \textbf{92.6} &  25.2 / 96.0 & 8.1 / 98.6 & \underline{32.9} / \textbf{93.7}\\
                SSD \cite{Sehwag2021} &  52.6 / 89.0 & 50.9 / 89.5 &  47.1 / 92.4 & 46.4 / 91.2 &  \textbf{13.1 / 97.8} & \textbf{0.9 / 99.8} & 35.1 / \underline{93.3}\\
            \midrule		
                \rowcolor[gray]{.95} EBM & \textbf{47.4} / \underline{89.9}&   \textbf{41.6} / \textbf{90.4}&  \textbf{32.7} / \textbf{93.9}&  \textbf{34.3} / \textbf{92.4}&  25.7 / 95.6&  5.7 / 98.9&  \textbf{31.3} / \textbf{93.5}\\
                \rowcolor[gray]{.95} ours & 50.0 / \underline{89.6}&   45.0 / \underline{90.2}&   40.7 / 93.1&   41.4 / \underline{91.9}&   \underline{16.4} / \underline{97.2}&   \underline{1.1} / \textbf{99.8}&   \textbf{32.4 / 93.7}\\
    		\bottomrule
    	\end{tabular}
    }
    \vspace{.5em}
    \caption{\textbf{OOD detection results on CIFAR-10.}}
    \label{supp:fig:ood_results_cifar10}
\end{table}

\begin{table}[ht]
\huge
\vspace{15mm}
    \centering
	\resizebox{\linewidth}{!}{%
    	\begin{tabular}{l  c c  c c  c c  c}
    		\toprule
       &  \multicolumn{2}{c}{\textit{Near-OOD}} & \multicolumn{2}{c}{\textit{Mid-OOD}}  & \multicolumn{2}{c}{\textit{Far-OOD}} & \\
  		\textbf{Method} & \textbf{C-10} & \textbf{TinyIN} & \textbf{LSUN} & \textbf{Places365} & \textbf{Textures} & \textbf{SVHN} & \textbf{Average}\\
        &  \Large{\text{FPR95$\downarrow$ / AUC $\uparrow$}} &\Large{\text{FPR95$\downarrow$ / AUC $\uparrow$}} & \Large{\text{FPR95$\downarrow$ / AUC $\uparrow$}} & \Large{\text{FPR95$\downarrow$ / AUC $\uparrow$}} & \Large{\text{FPR95$\downarrow$ / AUC $\uparrow$}} & \Large{\text{FPR95$\downarrow$ / AUC $\uparrow$}} & \Large{\text{FPR95$\downarrow$ / AUC $\uparrow$}}\\
    		\midrule
                MSP \cite{hendrycks17baseline}& \textbf{80.0} / \textbf{76.6} & 78.3 / 77.6 & \textbf{83.5} / 74.7 & \textbf{81.0} / \textbf{76.4} & 72.1 / 81.0 & 62.0 / 86.4 & 76.1 / 78.8\\
    		ODIN \cite{odin2018} & 81.4 / 76.4 & 78.7 / 76.2 & 86.1 / 72.0 & 82.6 / 74.5 & 62.4 / 85.2 & 80.7 / 80.4 & 78.6 / 77.5 \\
                Energy-Logits \cite{Liu2020} & \textbf{80.6} / \textbf{76.9} & 79.4 / 76.5 & 87.6 / 71.7 & 83.1 / 74.7 & 62.4 / 85.2 & 53.0 / 88.9 &  74.3 / 79.0 \\
                KNN \cite{Sun2022} & \underline{81.0} / \underline{76.1} & \textbf{76.0} / \textbf{81.1}  & 89.5 / \textbf{75.7}  & 82.1 / \textbf{75.7}  & 58.2 / 88.0  & 37.2 / 93.6 &  70.7 / \textbf{82.0} \\
                SSD \cite{Sehwag2021}& 85.6 / 73.6 & 82.7 / 77.0  & 87.8 / 73.8  & 84.6 / 74.5  & 36.6 / \textbf{92.4}  & \underline{19.9} / \textbf{96.4}  &  \underline{66.2} / \underline{81.3} \\
                \midrule
                \rowcolor[gray]{.95}EBM &  83.7 / 72.6&  78.1 / 78.8&  87.6 / 72.6&  83.0 / 74.2&  73.7 / 84.8&  52.2 / 90.4&  76.4 / 78.9\\
                \rowcolor[gray]{.95} ours & 85.0 / 74.2&  80.8 / 78.3&  86.9 / 74.2&  83.4 / 75.3&  \textbf{35.6 / 92.6}&  \textbf{18.3 / 96.6}&  \textbf{65.0 / 81.9}\\
    		\bottomrule
    	\end{tabular}
    }
    \vspace{.5em} 
    \caption{\textbf{OOD detection results on CIFAR-100.}}
    \label{supp:fig:ood_results_cifar100}
\end{table}

\end{document}